\documentclass[conference]{IEEEtran}
\usepackage{blindtext, graphicx}
\usepackage[font=scriptsize,justification=centering]{caption}
\usepackage{float}
\usepackage{mathtools}
\usepackage[bottom]{footmisc}
\usepackage{lipsum}
\usepackage{graphicx}

\usepackage{verbatim}
\usepackage{natbib}
\usepackage[utf8]{inputenc}
\usepackage[english]{babel}
\usepackage{amsthm}
\usepackage[linesnumbered,ruled]{algorithm2e}
\usepackage{algpseudocode}
\usepackage{wasysym}

\usepackage{array}
\newcolumntype{P}[1]{>{\centering\arraybackslash}p{#1}}

\usepackage{commath}
\def\tablename{Table}
\ifCLASSINFOpdf
\else
\fi
\begin{document}
%
\title{Neural Image Captioning}

\author{\IEEEauthorblockN{Lakshay Sharma}
\IEEEauthorblockA{Courant Institute of Mathematical Sciences\\
New York University\\
Email: ls4170@nyu.edu}
\and
\IEEEauthorblockN{Elaina Tan}
\IEEEauthorblockA{Courant Institute of Mathematical Sciences\\
New York University\\
Email: et1503@nyu.edu}}


%


\maketitle

\begin{abstract}
In recent years, the biggest advances in major Computer Vision tasks, such as object recognition, handwritten-digit identification, facial recognition, and many others., have all come through the use of Convolutional Neural Networks (CNNs). Similarly, in the domain of Natural Language Processing, Recurrent Neural Networks (RNNs), and Long Short Term Memory networks (LSTMs) in particular, have been crucial to some of the biggest breakthroughs in performance for tasks such as machine translation, part-of-speech tagging, sentiment analysis, and many others. These individual advances have greatly benefited tasks even at the intersection of NLP and Computer Vision, and inspired by this success, we studied some existing neural image captioning models that have proven to work well. In this work, we study some existing captioning models that provide near state-of-the-art performances, and try to enhance one such model. We also present a simple image captioning model that makes use of a CNN, an LSTM, and the beam search\footnote{https://en.wikipedia.org/wiki/Beam\_search} algorithm, and study its performance based on various qualitative and quantitative metrics.
\end{abstract}


%
\IEEEpeerreviewmaketitle

\section{Introduction}
 The problem of image captioning is at the intersection of Computer Vision and Natural Language Processing. Given an image, the model needs to generate a sensible, accurate, and grammatically correct caption for the image. The ability to generate coherent descriptions has utility in image classification and annotation for large sets of data, particularly with regards to image search engines. Additionally, image captioning can act as a feeder to further tasks such as video classification and image and text inference. This project aims to develop a model capable of this task and explore and quantify the results.

The report is outlined as follows. Section 2 gives a brief overview of related work. Section 3 gives an overview of the model architecture used. Section 4 discusses the methods used for experimentation and Section 5 discusses the results and analyzes the errors observed. Section 6 explores the robustness and generalizability of our model. Finally, Section 7 summarizes the findings from this report.

\section{Related Work}\label{sec:related}

The successes achieved by the use of neural networks in the fields of Computer Vision and Natural Language Processing individually have been reflected in breakthroughs in image captioning and scene understanding tasks. 

All of the best performing models in the recent past employ the same basic processing pipeline: using a pre-trained CNN to generate features for an input image, which is then fed to an RNN. For  a caption of $n$ words with each word being considered a state, at state $t$, the RNN receives as input a hidden state and the previously generated word/token. Each word is represented in the form of a \textit{word embedding} of a fixed dimension. At state $0$, the hidden state is the image and the previous token is a designated \textit{START} token. At training time, the RNN receives as input a hidden state and the last generated word. Therefore, the word generated as state $t$ is conditioned explicitly on the previous word/state $t-1$, and implicitly on the the entire sentence prior to state $t$ and the image features.

 \textit{Deep Visual-Semantic Alignments for Generating Image Descriptions} (\cite{Karpathy-Li}): This paper broke down what the authors believed were the necessary  components for generating high quality image captions. The focus of this work was to generate annotated regions that were represented with an embedding that, once combined with an RNN, was able to generate a full sentence describing the image. The takeaway was the importance of good embeddings as it inherently correlates with the quality of the sentence created. In our own models, we followed this example and focused on the features and captions related to the image as opposed to the image directly.

 \textit{Show, Attend and Tell: Neural Image Caption
Generation with Visual Attention} (\cite{ShowAttendTell}): This paper utilized feature, language, and attention inputs to build their model for captioning. Attention deconstructs the image into weighted sections that represent that section's supposed importance or relevance. Instead of weighing all features in the image equally, features that fall under regions with higher attention will be weighted higher in the caption generation, causing the caption to be more biased towards features found in areas where the attention was defined. 

This paper included an in-depth analysis on the difference in performance given the quality of the attention model. Although a good attention transformation can significantly improve the quality of the captions created, there is also a risk in incorporating it as an incorrect attention will cause bad captioning regardless of the captioning model's quality. This is why we did not utilize attention in our model from scratch, although it is part of the model we enhanced.

 \textit{Bottom-Up and Top-Down Attention for Image Captioning
and Visual Question Answering} (\cite{TopDown}): This paper built on the principles presented in Show, Attend, and Tell but produced a novel model that worked with the inputs in a different manner. This paper produced one of the best results in image captioning when compared to existing models and is the model we decided to focus on understanding and building upon. Top Down is also a relatively simple model by design that gains most of its power from the structure of the model. We decided to build upon this model in an attempt to create better or comparable results.

\section{Model Architecture}

We experimented with two different approaches to the model architecture: one where we analyzed the current state-of-the-art algorithms for image captioning and built an enhancement to an existing model, and one where we built our own model based on what we’ve learned from related works.
\raggedbottom
\subsection{LSTM-GRU}

For the model enhancement, which we will refer to as LSTM-GRU from here on out, we relied upon an existing implementation (\cite{Luo2017}). First, we tried to analyze and run the code to gauge its design and performance. Currently, the best models utilize three inputs: features, captions, and attention. For our purposes, these inputs were all extracted from existing feature detection models which take in the image as input. We used the idea that captions should be constructed, not necessarily directly from the images themselves, but from preprocessed inputs that have already simplified and restructured the image in a meaningful way. 

As one of the best models currently for image captioning is Top-Down, which uses beam search, attention, and an LSTM layer to generate captions, we chose this model for an attempted enhancement. From some preliminary investigation, we believed that a combination of a GRU cell and an LSTM cell would produce better results than the LSTM alone, without adversely affecting the time spent training. The rationale was that an LSTM uses three gates which allows it to better remember longer sequences, but does not train quickly and needs a large amount of data to learn. We hypothesized that adding a GRU layer following the LSTM layer would allow it to learn more early on while retaining the capabilities and benefits of long iteration learning from the LSTM layer (\cite{GRU}). Our enhanced model is composed of two layers of a single GRU and LSTM cell for both attention and feature vectors to help correlate them to the proposed caption. Beam search then calculates the quickest, best possible final caption that could be made from this model. 

\begin{figure}[H]
\centering
\includegraphics[width=3in]{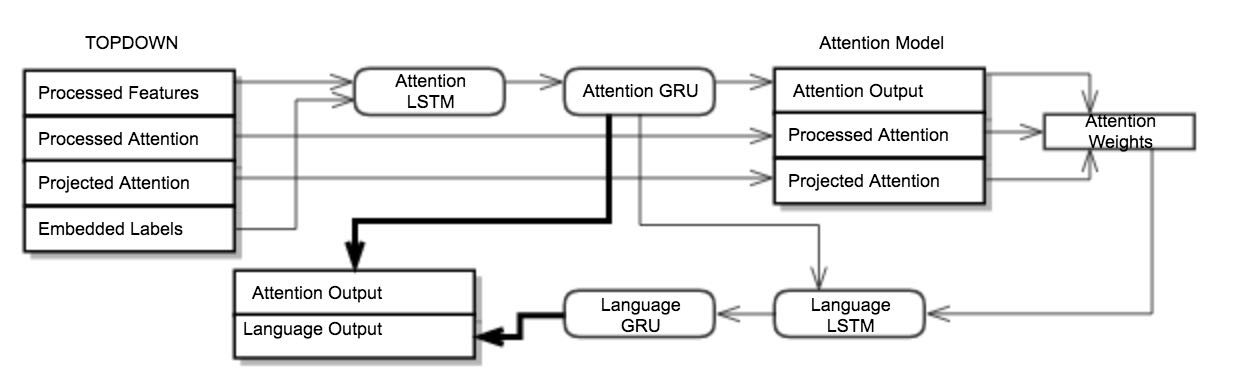}
\caption{LSTM-GRU model architecture}
\label{fig:lstmgru}
\end{figure}

The model is made up of two sets of LSTM-GRU layers, one for attention and one for language. The inputs to the model are the processed features, processed attention, projected attention, embedded labels, and state of the previous model. 

The attention LSTM takes in the concatenation of the previous hidden layer, the embedded labels, and the processed features. The hidden layer that is output is then used as the input to the attention GRU, which outputs a new attention hidden layer (referred to as attention output in the diagram and from here on out). Then the attention model is run on the concatenation of the attention output with the processed attention and projected attention to get the attention weights.  

The language LSTM takes in as input the concatenation of the attention weights and attention output, and then outputs a language hidden layer. This hidden layer is then used as input for the language GRU which outputs another hidden language layer (referred to as language output in the diagram and from here on out).

The final output from the model is the stacked attention output and language output.

\subsection{Specimen-Model}

Our experimental models were largely based on the same basic architecture followed by those described in Section \ref{sec:related}. 

An input image is first normalized and scaled, and then passed to a pre-trained object recognition CNN (example: \textit{Resnet152} (\cite{resnet})). Prior to this, the last layer of the CNN has been replaced by a fully-connected (FC) layer, that uses the output features generated at the CNN's penultimate layer to produce a feature vector of desired size. While the CNN is pre-trained and not tuned further, the last FC layer is untrained and its parameters are therefore optimized during training.

Considering each word to be a time step $t_{i}$, at time step $t_{0}$, the LSTM receives as input/hidden states the image embedding and the embedding for the \textit{$<$START$>$} token. At each further time step $t_{i}$, the hidden state from the previous step $h_{i-1}$ and the last token (from the ground truth caption at training time, or from the caption generated so far if the model is generating captions) are used as inputs, which are used to generate hidden state $h_{i}$, and the output feature vector $f_{i}$. $f_{i}$ is then passed to the classifier layer, which (using a softmax operation) produces a probability distribution over the entire vocabulary. The output word is considered to be the one with the highest probability. 

When generating models after training, at a given time step, instead of simply taking the word with the highest probability, beam search with a specified beam-size is applied to find the complete caption with the highest probability.

Figure \ref{fig:specimenmodel} describes this approach visually. 

Our model architecture (referred to as \textit{Specimen-Model} hereon) has the following fixed parameters:

\begin{tabular}{|r|l|}
  \hline
  Parameter & Value \\ \cline{2-2}
  \hline \hline
  Word embedding size & 256 \\
  LSTM hidden layer size & 256 \\
  Maximum caption length & 30 \\
  No. of training epochs & 10 \\
  Batch size & 32 \\
  Optimization & SGD \\
  Loss function & Cross-entropy \\
  Initial earning rate & 0.1 \\
  Momentum & 0.9 \\
  Weight decay & $1 \times 10^{-4}$ \\
  \hline
\end{tabular}

\begin{figure}[H]\label{fig:specimenmodel}
\centering
\includegraphics[width=3in]{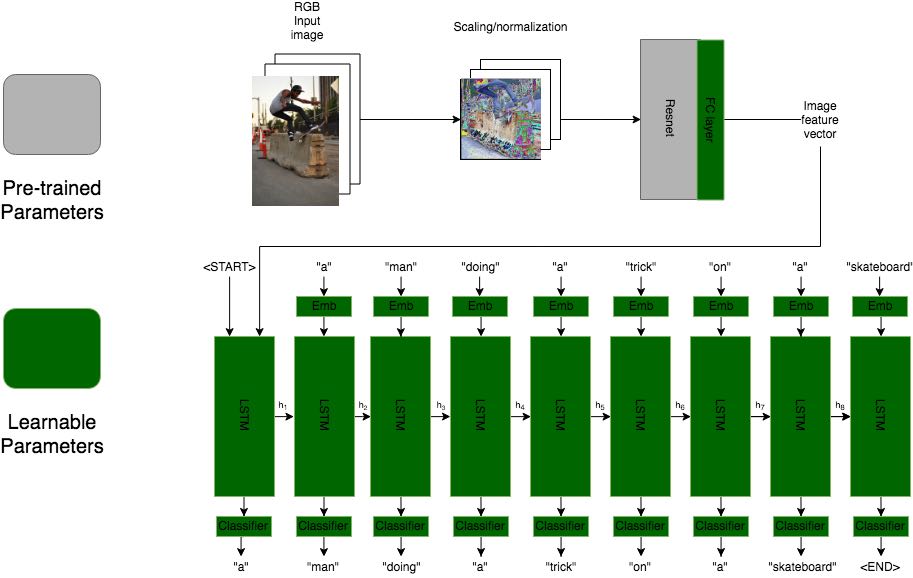}
\caption{Specimen-Model model architecture. Note: the multiple LSTM/classifier boxes shown represent the same layers at different time steps.}
\label{fig:specimenmodel}
\end{figure}

\section{Method}

\subsection{Dataset Used}
The dataset used was Microsoft COCO (Common Objects in Context) (\cite{lin2014microsoft}) Captioning Dataset (2014). It is comprised of 100,000+ image-caption pairs. There are 82,783 image-caption pairs in the training set, 40,504 image-caption pairs in the validation set, and 40,775 in the test set. We use the training and validation sets, which have 6,454,115 tokens, with 10,003 unique words.


\subsection{Preprocessing}

 Pre-trained \textit{Resnet50}, \textit{Resnet101}, and \textit{Resnet152} networks are used to generate attention and feature vectors for the images. The captions are also processed to be lower-case.



\subsection{Training}
In the Specimen-Model, we experimented with 1- and 2-layer LSTMs as well as different beam sizes. We also studied variations in results due to differences in the CNN generating the image features

For the LSTM-GRU model, we trained using features and attention generated from the following pre-trained models: Resnet50, Resnet101, and Resnet152. We trained for 25 epochs for each of the feature/attention sets used.

\section{Results}

\subsection{Qualitative results}
\begin{figure}[H]
\centering
\includegraphics[width=3.5in]{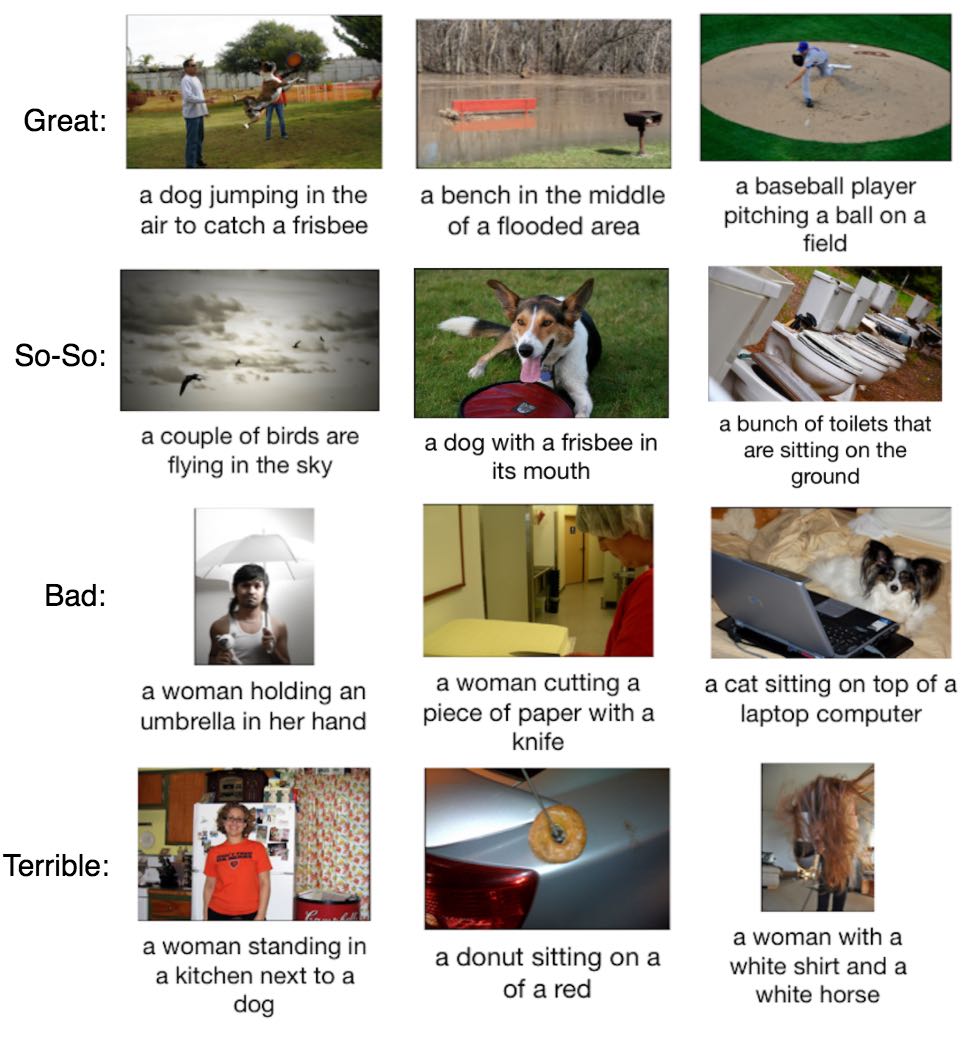}
\caption{Image captions generated by LSTM-GRU with Resnet50 and Specimen-Model with Resnet101}
\label{fig:results}
\end{figure}
\raggedbottom
 Figure \ref{fig:results} depicts a sample set of captions that were generated by our models, which we have attempted to sort qualitatively. 
 
 Above, we declare a great caption to be one that is an an accurate and illustrative description of what is happening in the picture. For example, LSTM-GRU is able to describe a flooded area instead of a lake as well as differentiate between a dog that is jumping versus standing or sitting. 
 
 A so-so caption is a description close to the ground truth (i.e. what is really happening in the picture) but missing or misreading intricacies, or with incorrect grammatical/semantic constructs. The first image for so-so captions describes a couple of birds as opposed to many; in the second, the Specimen-Model describes a dog with a frisbee in its mouth as opposed to in front of it; in the last, the model describes the toilets to be sitting on the ground, whereas a human would never describe a toilet to be sitting. 
 
 We label a bad caption to be an inaccurate but somewhat understandable caption of the image. Bad captions include mistakes such as gender, object misclassification, or relational errors. 
 
 A terrible caption contains one or more errors that can lead to complete misrepresentation of what is happening in the picture or the inability to form a complete caption. In the first image for terrible captions, LSTM-GRU describes a woman standing next to a dog, but there is no dog in the image. In the second image, the model is unable to even form a complete caption. 

\subsection{LSTM-GRU results}

For this (and other models), we evaluated performance using various scoring metrics on the MS-COCO validation set, as shown below.

\begin{table}[h]
\centering
\begin{tabular}{|P{1.1cm}|P{0.7cm}|P{0.7cm}|P{0.7cm}|P{0.7cm}|P{1cm}|P{0.8cm}|}
  \hline
 \textbf{Feature generator} & \textbf{Bleu-1}  & \textbf{Bleu-2} & \textbf{Bleu-3} & \textbf{Bleu-4} & \textbf{METEOR} & \textbf{CIDEr}  \\
\hline
  \textbf{Resnet50} & \textbf{0.798} & \textbf{0.650} & \textbf{0.515} & \textbf{0.406} & \textbf{0.293} & \textbf{1.249} \\  \hline
  \textbf{Resnet101} & 0.781 & 0.627 & 0.491 & 0.381 & 0.284 & 1.180\\
  \hline
  \textbf{Resnet152} & 0.784 & 0.631 & 0.496 & 0.387 & 0.287 & 1.201 \\
  \hline
\end{tabular}
\caption{LSTM-GRU scores after 25 epochs\label{tab:blocksizecomp}}
\label{table:lstmgru}
\end{table}

As depicted in Table \ref{table:lstmgru}, the enhancements made to the Top-Down model surpass the original model's existing scores, improving its best CIDEr result from 117.9 to 124.9 and Bleu-4 from 36.9 to 40.6\cite{TopDown}.

\subsection{Specimen-Model results}

Variations in output caption scores were observed while varying beam size, feature generator, and the number of LSTM layers. These are shown in Tables \ref{smr1}, \ref{smr2}, \ref{smr3}, and \ref{smr4}.

\begin{table}[h]
\centering
\begin{tabular}{|P{1.1cm}|P{0.7cm}|P{0.7cm}|P{0.7cm}|P{0.7cm}|P{1cm}|P{0.8cm}|}
  \hline
\textbf{Feature generator} & \textbf{Bleu-1}  & \textbf{Bleu-2} & \textbf{Bleu-3} & \textbf{Bleu-4} & \textbf{METEOR} & \textbf{CIDEr}  \\
\hline
  \textbf{Resnet50} & 0.655 & 0.\textbf{486} & 0.348 & 0.250 & 0.216 & 0.723 \\  \hline
  \textbf{Resnet101} & \textbf{0.666} & 0.466 & \textbf{0.350} & \textbf{0.252} & 0.215 & \textbf{0.728}\\
  \hline
  \textbf{Resnet152} & 0.659 & 0.481 & 0.345 & 0.249 & \textbf{0.218} & 0.726 \\
  \hline
\end{tabular}
\caption{Specimen-Model: 2 LSTM layers, beam size 3\label{smr1}}
\end{table}

\begin{table}[h]
\centering
\begin{tabular}{|P{1.1cm}|P{0.7cm}|P{0.7cm}|P{0.7cm}|P{0.7cm}|P{1cm}|P{0.8cm}|}
  \hline
\textbf{Feature generator} & \textbf{Bleu-1}  & \textbf{Bleu-2} & \textbf{Bleu-3} & \textbf{Bleu-4} & \textbf{METEOR} & \textbf{CIDEr}  \\
\hline
  \textbf{Resnet50} & 0.661 & 0.484 & 0.349 & 0.253 & 0.215 & 0.726 \\  \hline
  \textbf{Resnet101} & \textbf{0.663} & \textbf{0.486} & \textbf{0.351} & 0.255 & 0.214 & 0.731\\
  \hline
  \textbf{Resnet152} & \textbf{0.663} & \textbf{0.486} & \textbf{0.351} & \textbf{0.256} & \textbf{0.218} & \textbf{0.733} \\
  \hline
\end{tabular}
\caption{Specimen-Model: 2 LSTM layers, beam size 4\label{smr2}}
\end{table}

\begin{table}[h]
\centering
\begin{tabular}{|P{1.1cm}|P{0.7cm}|P{0.7cm}|P{0.7cm}|P{0.7cm}|P{1cm}|P{0.8cm}|}
  \hline
\textbf{Feature generator} & \textbf{Bleu-1}  & \textbf{Bleu-2} & \textbf{Bleu-3} & \textbf{Bleu-4} & \textbf{METEOR} & \textbf{CIDEr}  \\
\hline
  \textbf{Resnet50} & 0.655 & 0.474 & 0.338 & 0.242 & 0.214 & 0.703 \\  \hline
  \textbf{Resnet101} & 0.660 & 0.482 & 0.345 & 0.247 & 0.214 & 0.713\\
  \hline
  \textbf{Resnet152} & \textbf{0.668} & \textbf{0.490} & \textbf{0.352} & \textbf{0.254} & \textbf{0.218} & \textbf{0.740} \\
  \hline
\end{tabular}
\caption{Specimen-Model: 1 LSTM layer, beam size 3\label{smr3}}
\end{table}

\begin{table}[!htbp]
\centering
\begin{tabular}{|P{1.1cm}|P{0.7cm}|P{0.7cm}|P{0.7cm}|P{0.7cm}|P{1cm}|P{0.8cm}|}
  \hline
\textbf{Feature generator} & \textbf{Bleu-1}  & \textbf{Bleu-2} & \textbf{Bleu-3} & \textbf{Bleu-4} & \textbf{METEOR} & \textbf{CIDEr}  \\
\hline
  \textbf{Resnet50} & 0.652 & 0.473 & 0.339 & 0.245 & 0.213 & 0.703 \\  \hline
  \textbf{Resnet101} & 0.658 & 0.482 & 0.346 & 0.251 & 0.214 & 0.718\\
  \hline
  \textbf{Resnet152} & \textbf{0.662} & \textbf{0.486} & \textbf{0.351} & \textbf{0.255} & \textbf{0.217} & \textbf{0.737} \\
  \hline
\end{tabular}
\caption{Specimen-Model: 1 LSTM layer, beam size 4\label{smr4}}
\end{table}

As the results show, there is almost always a slight improvement in scores when the beam size changes from 3 to 4. An interesting thing to note however is that having a 2-layer LSTM versus having a 1-layer LSTM does not guarantee an increase in scores. While the CIDEr scores for the Specimen-Model with 2 LSTM layers using Resnet50 and Resnet101 are greater than the CIDEr scores for the Specimen-Model with 1 LSTM layer, when using Resnet152 the 1-layer model does better than the 2-layer model.

\begin{figure}[H]
\centering
\includegraphics[width=2in]{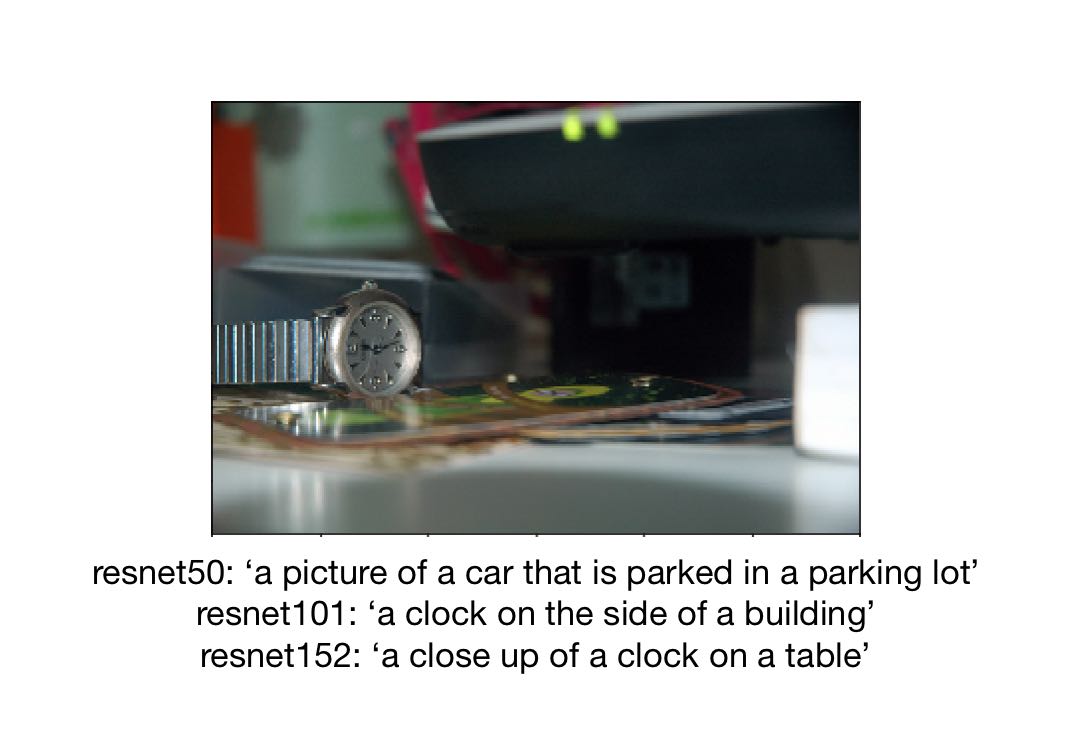}
\caption{Variations in captions generated by Specimen-Model based on different image feature generators}
\label{fig:rescomp1}
\end{figure}

\begin{figure}[H]
\centering
\includegraphics[width=2in]{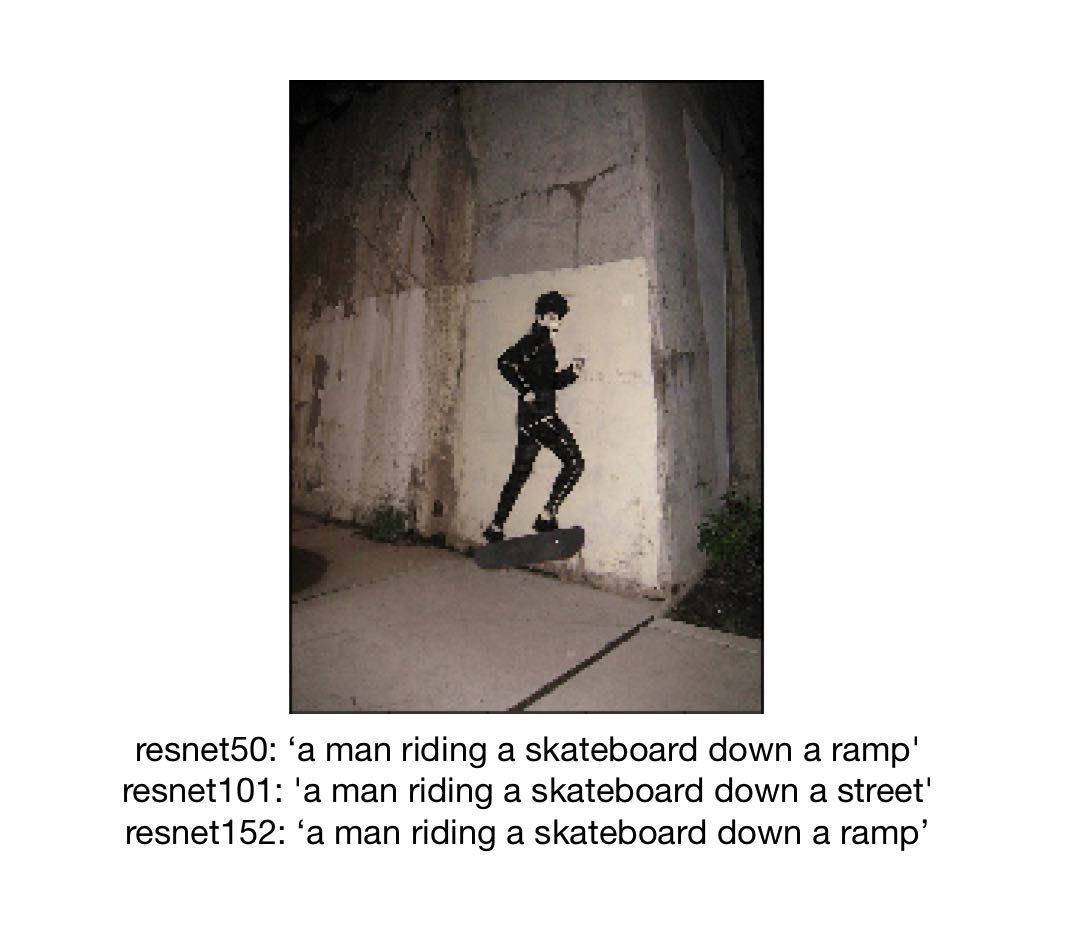}
\caption{Variations in captions generated by Specimen-Model based on different image feature generators}
\label{fig:rescomp2}
\end{figure}

\begin{figure}[H]
\centering
\includegraphics[width=2in]{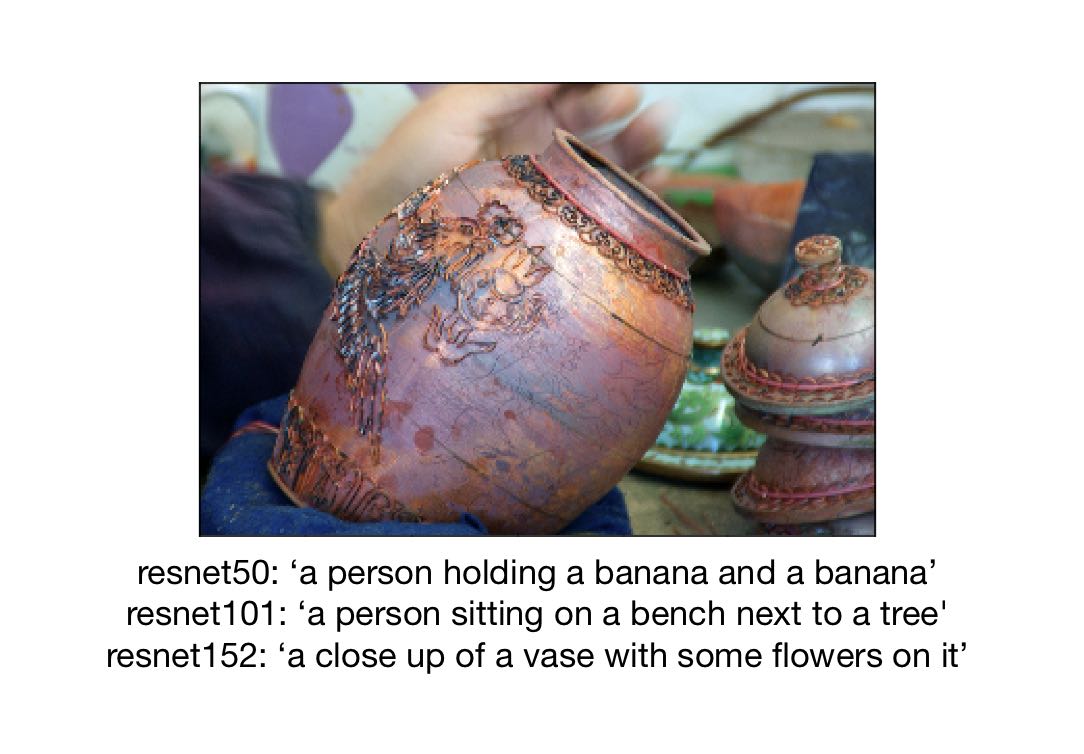}
\caption{Variations in captions generated by Specimen-Model based on different image feature generators}
\label{fig:rescomp3}
\end{figure}

As expected, a deeper feature generator CNN almost always produces superior results. However, some validation instances are observed where a shallower CNN produces a more correct output caption than a deeper CNN. Some of these results have been shown in figures \ref{fig:rescomp1}, \ref{fig:rescomp2}, and \ref{fig:rescomp3}.

\subsection{Error Analysis}

The captions generated, when incorrect, tended to exhibit errors that broadly fell within the following categories: counting error, gender error, existence error, relational error, color error, and classification error. These errors are defined and summarized below.

\begin{itemize}
    \item Counting Error: when the model is not able to determine the correct number of objects, animals, or persons in an image. 
    
\begin{minipage}{\linewidth}
\begin{figure}[H]
\centering
\includegraphics[width=1.5in]{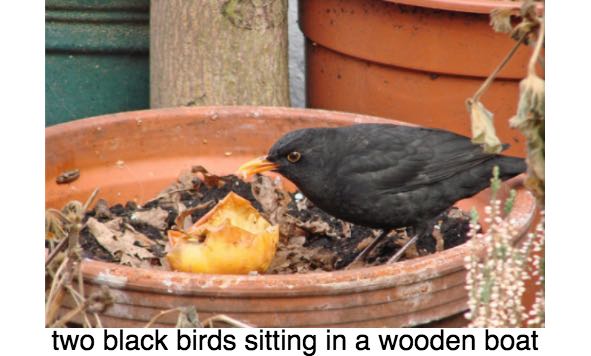}
\caption{A combination of counting and classification error. An image caption that incorrectly counts two birds when there is only one (LSTM-GRU)}
\label{fig:countingerror}
\end{figure}

    \item Gender Error: when the model is not able to determine the correct gender of persons in an image.

\begin{figure}[H]
\centering
\includegraphics[width=1.5in]{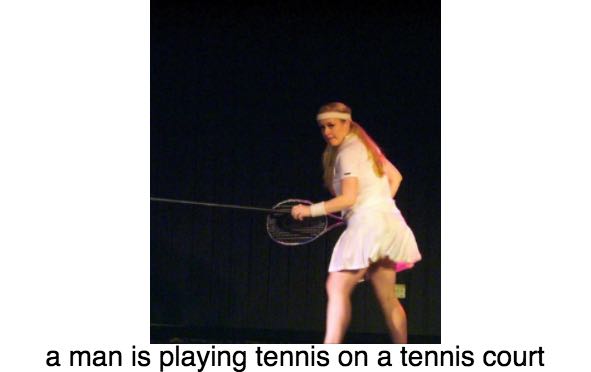}
\caption{An image caption that incorrectly identifies a man instead of a woman (Specimen-Model)}
\label{fig:gendererror}
\end{figure}

    \item Existence Error: when the model is unable to correctly determine whether or not an object, animal, or person exists in an image.

\begin{figure}[H]
\centering
\includegraphics[width=1.5in]{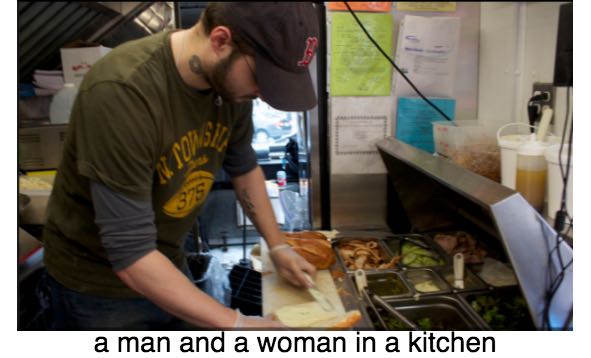}
\caption{An image caption that incorrectly identifies a man and a woman in a kitchen but there is only a man (Specimen-Model)}
\label{fig:existenceerror}
\end{figure}
    
    \item Color Error: when the model is unable to determine the correct color of an object, animal, or person in an image.

\begin{figure}[H]
\centering
\includegraphics[width=1.8in]{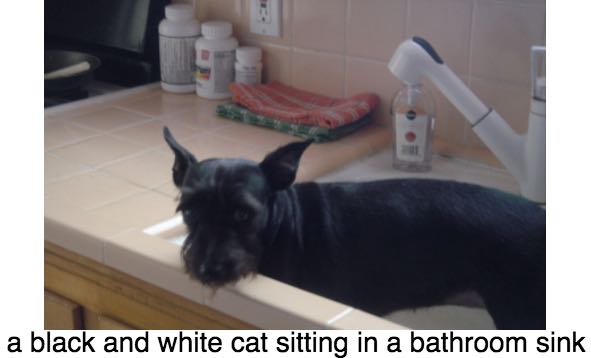}
\caption{A combination of color and classification errors. An image caption that incorrectly identifies a black and white cat when the animal is a black dog (Specimen-Model)}
\label{fig:colorerror}
\end{figure}

    \item Classification Error: when the model incorrectly identifies an object, animal, or person in an image.

\begin{figure}[H]
\centering
\includegraphics[width=1.5in]{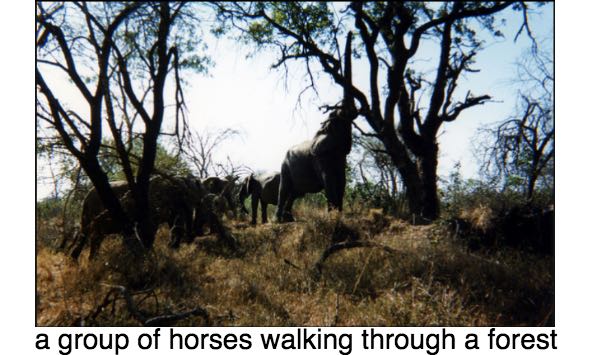}
\caption{An image caption that incorrectly identifies elephants as horses (LSTM-GRU)}
\label{fig:classificationerror}
\end{figure}
\end{minipage}
\pagebreak
\begin{minipage}{\linewidth}

\item Relational Error: when the model is unable to correctly determine an action or contextual relationship between objects, animals, or persons in an image.
\begin{figure}[H]
\centering
\includegraphics[width=1.5in]{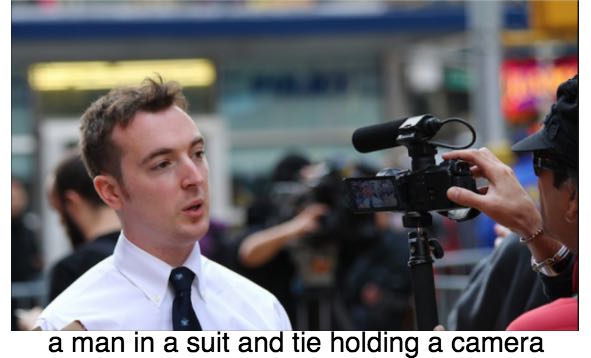}
\caption{An image caption that incorrectly identifies the man in the suit and tie as holding the camera instead of being filmed by the camera (LSTM-GRU)}
\label{fig:relationalerror}
\end{figure}
\end{minipage}
\end{itemize}

As can be seen from the earlier examples, some of these errors can be minor and lead only to partially incorrect captions. However, certain errors such as existence, classification, and relational, or combinations of these errors, can lead to contradictions or nonsensical statements made by the models when attempting to caption the images.

\section{Model robustness/generalizability}

For Specimen-Model, we try a nearest-neighbors test to see how well the model functions and adjust to deviations.

Given some image $i$ and its output feature vector $v \in R^{256}$  generated by the trained FC layer of the CNN, a caption is generated by the model. Each word $w_{i}$ in this generated caption has a corresponding word embedding in the embedding matrix of the model. For the purpose of this study, we consider a caption to be represented by its embedding $c$, which is the average of the embeddings of all the words $w_{i},...,w_{n}$ that constitute the caption.

Even though $v$ and $c$ don't exist in the same embedding space, they could be considered counterparts. 

1000 samples from the COCO validation set were randomly selected, and image feature vectors were generated for each of these (as described above). From this, a random image $i$ and its feature vector $v$ were selected, and the $k$ (3 in our case) vectors that are most similar to $v$ (cosine similarity is used as the measure of similarity) were found. Let us call this set consisting of $v$ and its nearest-neighbors $S_{i}$.

For these same 1000 samples, the output captions were generated (using Specimen-Model), and  each caption was represented as an 'average embedding' as described above. Given some randomly chosen image $i$ and its generated feature vector $v$, we find the embedding $c$ of the corresponding caption that was generated. Similar to above, we find the 3 embeddings/vectors that are most similar to $c$. Let us call $c$ and its nearest-neighbors $S_{c}$.

If the model generalizes well, the images corresponding to the nearest neighbors of $v$ should be accurately describable by the nearest neighbors of $c$.

Note: the results and figures referred to below can be found in the Appendix.

Our findings showed that, in many cases, the pair $S_{i}$ and $S_{c}$ are such that each caption represented by $S_{c}$ can describe, with reasonable accuracy, each image represented by the vectors in $S_{i}$. In fact, certain cases are such that there is a correspondence in the k-nearest neighbors of $v$ and those of $c$, i.e. some image whose feature vector is among the 3 nearest vectors of $v$ also has its generated caption's vector/embedding among the 3 nearest vectors of $c$ (Fig. \ref{nn1}, Fig. \ref{nn2}, Fig. \ref{nn3}).

There were also some pairs $S_{i}$ and $S_{c}$ where, even though there were no such correspondences, the captions represented by $S_c$ (considered collectively) were fairly accurate in describing the images represented by $S_{i}$ (Fig. \ref{nn5}).

Another finding was that there were certain pairs $S_{i}$ and $S_{c}$ such that the k-nearest neighbors of $c$ were very bad descriptions for the images represented by $S_{i}$. However, in many of these, it was seen that the reference caption represented by $c$ was itself not a good description of the image represented by $v$ (Fig. \ref{nn4}). However, despite this, qualitative similarity within $S_{c}$ may suggest generalizability too. 

On the downside, it was seen that many sets of nearest-neighbor captions $S_{c}$ were not very diverse (and sometimes identical) in their vocabulary and/or sentence structure (Fig. \ref{nn4}). This suggests the model overfits the language style of the dataset.







\section{Conclusion}
\begin{itemize}
    \item This work begins with a brief outline of related works in the field of image captioning.
    \item We proposed two model architectures: one an enhancement on a current state-of-the-art implementation and one a simple CNN/LSTM model that demonstrated the ability of a generic model to generate coherent image captions.
    \item Experiments were performed, and performance variations were quantified with variations in beam search, input features generated, and the number of LSTM layers. This was followed by a k-nearest-neighbors robustness/generalizability check. 
    \item The errors generated by the model were sorted into definitive categories.
    \item For future work, we would like to extend the relatively simple Specimen-Model to use image/text localization techniques such as attention with the aim of making its performance more comparable with the state-of-the-art captioning models.
\end{itemize}

\section{Acknowledgments}
    We would like to thank Rob Fergus for his guidance through the course of this work, NYU High Performance Computing and Erica Golin for support in terms of hardware resources, and \cite{Luo2017} and \cite{captionGen} for making their code publicly available.

\pagebreak

\onecolumn

\appendix[]

An image with a colored border and a caption with the same font color represent a corresponding image-caption pair (i.e., this caption was the one generated by the model for this image).

\begin{figure}[H]
    \centering
    \onecolumn\includegraphics[width=7.25in]{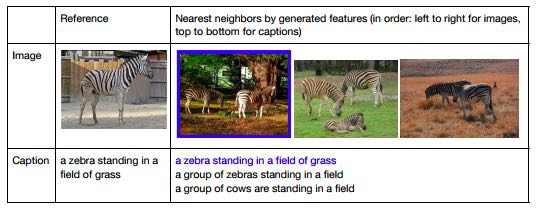}
    \caption{Nearest-neighbors result (1)}
    \label{nn1}
    \captionsetup{justification=centering}
    \label{fig:existenceerror}
\end{figure}

\begin{figure}[H]
    \centering
    \onecolumn\includegraphics[width=7.25in]{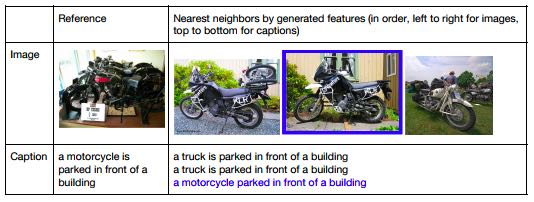}
    \caption{Nearest-neighbors result (2)}
    \label{nn2}
    \captionsetup{justification=centering}
    \label{fig:existenceerror}
\end{figure}

\begin{figure}[H]
    \centering
    \onecolumn\includegraphics[width=7.25in]{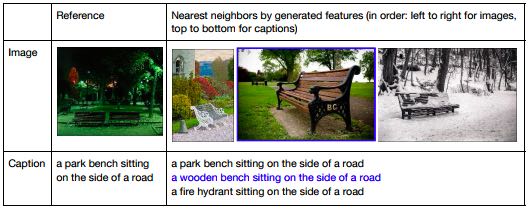}
    \caption{Nearest-neighbors result (3)}
    \label{nn3}
    \captionsetup{justification=centering}
    \label{fig:existenceerror}
\end{figure}

\begin{figure}[H]
    \centering
    \onecolumn\includegraphics[width=7.25in]{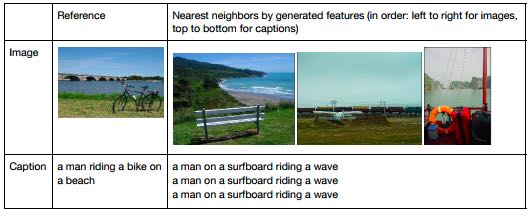}
    \caption{Nearest-neighbors result (5)}
    \label{nn4}
    \captionsetup{justification=centering}
    \label{fig:existenceerror}
\end{figure}

\begin{figure}[H]
    \centering
    \onecolumn\includegraphics[width=7.25in]{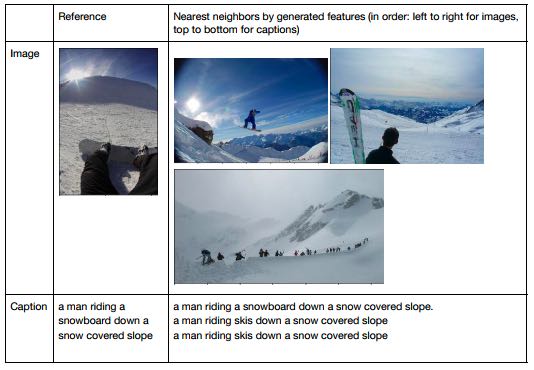}
    \caption{Nearest-neighbors result (5)}
    \label{nn5}
    \captionsetup{justification=centering}
    \label{fig:existenceerror}
\end{figure}

\ifCLASSOPTIONcaptionsoff
  \newpage
\fi



%

%
\bibliographystyle{abbrvnat}
\setcitestyle{authoryear,open={((},close={))}}
\bibliography{references}




\end{document}